\documentclass[letterpaper]{article} 
\usepackage{aaai23_arxiv}  
\usepackage{times}  
\usepackage{helvet}  
\usepackage{courier}  
\usepackage[hyphens]{url}  
\usepackage{graphicx} 
\usepackage{natbib}  
\usepackage{caption} 
\usepackage{algorithm}
\usepackage{algorithmic}

\usepackage{newfloat}
\usepackage{listings}
\DeclareCaptionStyle{ruled}{labelfont=normalfont,labelsep=colon,strut=off} 
\floatstyle{ruled}
\newfloat{listing}{tb}{lst}{}
\floatname{listing}{Listing}
\usepackage{amsmath}
\usepackage{amssymb}
\usepackage{mathtools}
\usepackage{amsthm}

\usepackage[capitalize,noabbrev]{cleveref}

\theoremstyle{plain}
\newtheorem{theorem}{Theorem}[section]
\newtheorem{proposition}[theorem]{Proposition}

\theoremstyle{definition}
\newtheorem{definition}[theorem]{Definition}

\theoremstyle{remark}

\usepackage{comment}
\usepackage{makecell}
\usepackage[labelformat=simple]{subcaption}

\usepackage{enumerate}

\newcommand{\req}[1]{Eq.~(\ref{#1})}
\newcommand{\rfig}[1]{Fig.~\ref{#1}}
\newcommand{\rtab}[1]{Tab.~\ref{#1}}
\newcommand{\rsec}[1]{Section~\ref{#1}}

\title{Fast Saturating Gate for Learning Long Time Scales with \\ Recurrent Neural Networks}

\author {
    Kentaro Ohno, 
    Sekitoshi Kanai, 
    Yasutoshi Ida 
}
\affiliations {
    NTT \\ 
    \{kentaro.ohno.tf, sekitoshi.kanai.fu,  yasutoshi.ida.yc\}@hco.ntt.co.jp
}

\begin{document}

\maketitle

\begin{abstract}
Gate functions in recurrent models, such as an LSTM and GRU, play a central role in learning various time scales in modeling time series data by using a bounded activation function.
However, it is difficult to train gates to capture extremely long time scales due to gradient vanishing of the bounded function for large inputs, which is known as the saturation problem.
We closely analyze the relation between saturation of the gate function and efficiency of the training.
We prove that the gradient vanishing of the gate function can be mitigated by accelerating the convergence of the saturating function, i.e., making the output of the function converge to 0 or 1 faster.
Based on the analysis results, we propose a gate function called fast gate that has a doubly exponential convergence rate with respect to inputs by simple function composition.
We empirically show that our method outperforms previous methods in accuracy and computational efficiency on benchmark tasks involving extremely long time scales.

\end{abstract}

\section{Introduction}

Recurrent neural networks (RNNs) are models suited to processing sequential data in various applications, e.g., speech recognition~\cite{ling2020deep} and video analysis~\cite{zhu2020faster}.
The most widely used RNNs are a long short-term memory (LSTM)~\cite{hochreiter1997long} and gated recurrent unit (GRU)~\cite{cho2014learning}, which has a gating mechanism.
The gating mechanism controls the information flow in the state of RNNs via multiplication with a gate function bounded to a range $[0,1]$.
For example, when the forget gate takes a value close to 1 (or 0 for the update gate in the GRU), the state preserves the previous information.
On the other hand, when it gets close to the other boundary, the RNN updates the state by the current input.
Thus, in order to represent long temporal dependencies of data involving hundreds or thousands of time steps,
it is crucial for the forget gate to take values near the boundaries~\cite{tallec2018can,mahto2020multi}.

However, it is difficult to train RNNs so that they have the gate values near the boundaries.
Previous studies hypothesized that this is due to gradient vanishing for the gate function called \emph{saturation}~\cite{chandar2019towards,gua2020improving}, i.e., the gradient of the 
gate function near the boundary is too small to effectively update the parameters.
To avoid the saturation problem, a previous study used unbounded activation functions~\cite{chandar2019towards}.
However, this makes training unstable due to the gradient explosion~\cite{pascanu2013difficulty}.
Another study introduced residual connection for a gate function to push the output value toward boundaries, hence mitigating the saturation problem~\cite{gua2020improving}.
However, it requires additional computational cost due to increasing the number of parameters for another gate function.
For broader application of gated RNNs, a more efficient solution is necessary.

\begin{figure}[t]
    \centering
    \includegraphics[width=0.34\paperwidth]{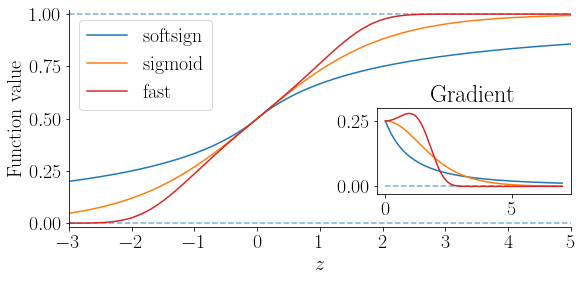}
    \caption{
        Function values and derivative of various bounded activation functions. 
        Sigmoid function $\sigma$ (orange) exponentially converges to 1 as $z\to \infty$, since $1- \sigma(z) = \frac{1}{1+e^z} \approx e^{-z}$.
        Derivative also decays exponentially.
        Normalized version of softsign function $\mathrm{softsign} (z) = \frac{z}{1 + |z|}$ (blue) converges to 1 more slowly.
        Fast gate (red) is proposed gate function.
        Although gradient decays faster than sigmoid function, it provably helps learning values near boundaries.
    }
    \label{fig:saturation}
\end{figure}

To overcome the difficulty of training, we propose a novel activation function for the forget gate based on the usual sigmoid function, which we call the \emph{fast gate}. Modification of the usual sigmoid gate to the fast gate is simple and easy to implement since it requires only one additional function composition.
To this end, we analyze the relation between the saturation and gradient vanishing of the bounded activation function.
Specifically, we focus on the convergence rate of the activation function to the boundary, which we call the \emph{order of saturation}.
For example, the sigmoid function $\sigma(z) = 1/ (1 + e^{-z})$ has the exponential order of saturation, i.e., $1-\sigma(z) = O(e^{-z})$ (see \rfig{fig:saturation}), and the derivative also decays to 0 exponentially as $z$ goes to infinity.
When a bounded activation function has a higher order of saturation, the derivative decays much faster as the input grows.
Since previous studies have assumed that the decaying derivative on the saturating regime causes the stuck of training~\cite{ioffe2015batch}, it seems that a higher order of saturation would lead to poor training.
Contrarily to this intuition, 
we prove that a higher order of saturation alleviates the gradient vanishing on the saturating regime
through observation on a toy problem for learning long time scales.
This result indicates that functions saturating superexponentially are more suitable for the forget gate to learn long time scales than the sigmoid function.
On the basis of this observation, we explore a method of realizing such functions by composing functions which increase faster than the identity function (e.g., $\alpha(z) = z + z^3$) as $\sigma ( \alpha(z) )$.
We find that the hyperbolic sinusoidal function is suitable for achieving a higher order of saturation in a simple way, and we obtain the fast gate.
Since the fast gate has a doubly exponential order of saturation $O(e^{-e^z})$, it improves the trainability of gated RNNs for long time scales of sequential data.
We evaluate the computational efficiency and accuracy of a model with the fast gate on several benchmark tasks, including synthetic tasks, pixel-by-pixel image classification, and language modeling, which involve a wide range of time scales.
The model with the fast gate empirically outperforms other models including an LSTM with the sigmoid gate and variants recently proposed for tackling the saturation problem~\cite{chandar2019towards,gua2020improving} in terms of accuracy and the convergence speed of training while maintaining stability of training.
Further visualization analysis of learning time scales shows that our theory fits the learning dynamics of actual models and that the fast gate can learn extremely long time scales of thousands of time steps.

Our major contributions are as follows:
\begin{itemize}
    \item We prove that gate functions which saturate faster actually \emph{accelerates} learning values near boundaries.
    The result indicates that fast saturation improves learnability of gated RNNs on data with long time scales.
    \item We propose the \emph{fast gate} that saturates faster than the sigmoid function.
    In spite of its simplicity, the fast gate achieves a doubly exponential order of saturation, and thus effectively improves learning of long time scales.
    \item We evaluate the effectiveness of the fast gate against recently proposed methods such as an NRU~\cite{chandar2019towards} and a refine gate~\cite{gua2020improving}. 
    The results verify that the fast gate robustly improves the learnability for long-term dependencies in both synthetic and real data.
\end{itemize}


\section{Preliminaries}\label{sec:preliminaries}

\subsection{Time Scales in Gated RNNs}\label{subsec:timescale}

In this section, we review gated RNNs and their time scale interpretation~\cite{tallec2018can}.
We begin with an LSTM~\cite{hochreiter1997long}, which is one of the most popular RNNs.
An LSTM has a memory cell $c_t \in \mathbb{R}^n$ and hidden state $h_t \in \mathbb{R}^n$ inside,
which are updated depending on the sequential input data $x_t$ at each time step $t = 1,2,\cdots$ by
\begin{align}\label{eq:lstm}
    c_t &= f_t \odot c_{t-1} + i_t \odot \tilde{c}_t  \\
    h_t &= o_t \odot \tanh{(c_t)}                     \\
    f_t &= \sigma (W_f x_t + U_f h_{t-1} + b_f)     \\
    i_t &= \sigma (W_i x_t + U_i h_{t-1} + b_i)     \\
    \tilde{c}_t &= \tanh (W_c x_t + U_c h_{t-1} + b_c)     \\
    o_t &= \sigma (W_o x_t + U_o h_{t-1} + b_o)
\end{align}
where $W_*, U_*$ and $b_*$ are weight and bias parameters for each $* \in \{f, i, c, o\}$.
The sigmoid function $\sigma$ is defined as
\begin{align} 
    \sigma(x) = \frac{1}{1+e^{-x}}.
\end{align}
$f_t, i_t, o_t \in [0,1]^n$ are called forget, input, and output gates, respectively.
They were initially motivated as a binary mechanism, i.e., switching on and off,
allowing information to pass through~\cite{gers1999learning}.
The forget gate has been reinterpreted
as the representation for time scales of memory cells~\cite{tallec2018can}.
Following that study, we simplify \req{eq:lstm}
by assuming $\tilde{c}_t = 0$ for an interval $t \in [t_0, t_1]$.
Then, we obtain
\begin{align}\label{eq:memory_decay}
    c_{t_1} &= f_{t_1} \odot c_{t_1-1}     \\
        &= \bar{f}^{t_1-t_0} \odot c_{t_1-t_0},
\end{align}
where $\bar{f} = (\prod_{s=t_0+1}^{t_1} f_s)^{\frac{1}{t_1-t_0}}$ is the (entry-wise) geometric mean of the values of the forget gate.
Through \req{eq:memory_decay}, the memory cell $c_t$ 
loses its information on data up to time $t_0$ exponentially, and the entry of $\bar f$ represents its (averaged) decay rate.
This indicates that, in order to capture long-term dependencies of the sequential data, the forget gate is desired to take values near 1 on average.
We refer the associated time constant\footnote{An exponential function $F(t) = e^{-\alpha t}$ of time $t$ decreases by a factor of $1/e$ in time $T = 1/\alpha$, which is called the time constant.}
$T = - 1/ \log {\bar f}$ as the \emph{time scale} of units,
which has been empirically shown to illustrate well the temporal behavior of LSTMs~\cite{mahto2020multi}.

The above argument applies not only to an LSTM, but also to general gated RNNs including a GRU~\cite{cho2014learning} with state update of the form
\begin{align}
    h_t = f_t \odot h_{t-1} + i_t \odot \tilde{h}_t,
\end{align}
where $h_t, f_t, i_t$ denotes the state, forget gate, and input gate, respectively,
and $\tilde{h}_t$ is the activation to represent new information at time $t$.
Here again, the forget gate $f_t$ takes a role to control the time scale of each unit of the state.

\subsection{Saturation in Gating Activation Functions}\label{subsec:saturation}

The sigmoid function $\sigma(z)$ in the gating mechanism requires large $z$ to take a value near 1 as the output.
On the other hand, the derivative $\sigma'(z)$ takes exponentially small values for $z \gg 0$ (\rfig{fig:saturation}).
Thus, when a gated model needs to learn large gate values such as $0.99$ with gradient methods,
parameters in the gate cannot be effectively updated due to gradient vanishing.
This is called \emph{saturation} of bounded activation functions~\cite{gulcehre2016noisy}.
The behavior of gate functions on the saturating regime is important for gated RNNs because
forget gate values need to be large to represent long time scales as explained in \rsec{subsec:timescale}.
That is, gated RNNs must face saturation of the forget gate to learn long time scales.
Thus, it is hypothesized that saturation causes
difficulty in training gated RNNs for data with extremely long time scales~\cite{chandar2019towards,gua2020improving}.

\section{Related Work}\label{sec:related_work}

Although there is abundant literature on learning long-term dependencies with RNNs, we outline the most related studies in this sectoin due to the space limitation and provide additional discussion of other studies in Appendix~\ref{app:related_work}.

Several studies investigate the time scale representation of the forget gate function to improve learning on data involving long-term dependencies~\cite{tallec2018can,mahto2020multi}.
For example, performance of LSTM language models can be improved by fixing the bias parameter of the forget gate in accordance with a power law of time scale distribution, which underlies natural language~\cite{mahto2020multi}.
Such techniques require us to know the appropriate time scales of data a priori, which is often difficult.
Note that this approach can be combined with our method since it is complementary with our work.

Several modifications of the gate function have been proposed to tackle the saturation problem.
The noisy gradient for a piece-wise linear gate function was proposed to prevent the gradient to take zero values~\cite{gulcehre2016noisy}.
This training protocol includes hyperparameters controlling noise level, which requires manual tuning.
Furthermore, such a stochastic approach can result in unstable training due to gradient estimation bias \cite{bengio2013estimating}.
The refine gate~\cite{gua2020improving} was proposed as another modification introducing a residual connection to push the gate value to the boundaries.
It is rather heuristic and does not provide theoretical justification.
It also requires additional parameters for the auxiliary gate, which increases the computational cost for both inference and training.
In contrast, our method theoretically improves learnability and does not introduce any additional parameters.
Another study suggests that omitting gates other than the forget gate makes training of models for long time scales easier~\cite{van2018unreasonable}.
However, such simplification may lose the expressive power of the model and limit its application fields.
\citet{chandar2019towards} proposed an RNN with a non-saturating activation function to directly avoid the gradient vanishing due to saturation.
Since its state and memory vector evolves in unbounded regions, the behavior of the gradient can be unstable depending on tasks.
Our method mitigates the gradient vanishing by controlling the order of saturation, while maintaining the bounded state transition.

\section{Analysis on Saturation and Learnability}\label{sec:saturation_learnability}

We discuss the learning behavior of the forget gate for long time scales.
First, we formulate a problem of learning long time scales in a simplified setting.
Next, we relate the efficiency of learning on the problem to the saturation of the gate functions.
We conclude that the faster saturation makes learning more efficient.
All proofs for mathematical results below are given in Appendix~\ref{app:proofs}.

\subsection{Problem Setting}\label{subsec:toy-problem}

Recall \req{eq:memory_decay}, which describes the time scales of the memory cell $c_t$ of an LSTM via exponential decay.
Let the memory cell at time $t_1$ be $c_{t_1} = \lambda c_{t_0}$ with  $\lambda \in [0,1]$.
Requiring long time scales corresponds to getting $\lambda$ close to 1.
Therefore, we can consider a long-time-scale learning problem as minimizing a 
loss function $L$ that measures discrepancy of $c_{t_1}$ and $\lambda_* c_{t_0}$ where $\lambda_* \in [0,1]$ is a desired value close to 1.
We take $L$ as the absolute loss for example.
Then, we obtain
\begin{align}
    L
        &= |c_{t_1} - \lambda_* c_{t_0}|  \\
        &= |\bar f^{t_1 - t_0} c_{t_0} - \lambda_* c_{t_0}|   \\
        &= c_{t_0} |\bar f^{t_1 - t_0} - \lambda_*|,
\end{align}
using \req{eq:memory_decay}.
Let $z_t = W_f x_t + U_f h_{t-1} + b_f$, so that $f_t = \sigma(z_t)$.
Since we are interested in the averaged value of $f_t$, we consider $z_t$ to be time-independent, that is, $z_t = z$ in the same way as \citet{tallec2018can}.
The problem is then reduced to a problem to obtain $z$ that minimizes
\begin{align}\label{eq:toy-problem}
    L(z)
        = c_{t_0} |\sigma(z)^{t_1 - t_0} - \lambda_*|.
\end{align}
We consider this as the minimal problem to analyze the learnability of the forget gate for long time scales.
Note that since the product $c_{t_1} = \bar{f}^{t_1-t_0} c_{t_0}$ is taken element-wise, we can consider this as a one-dimensional problem.
Furthermore, the global solution can be explicitly written as $z = \sigma^{-1}(\lambda_*^{1/(t_1-t_0)})$ where $\sigma^{-1}$ is an inverse of $\sigma$.

Next, we consider the learning dynamics of the model on the aforementioned problem \req{eq:toy-problem}.
RNNs are usually trained with gradient methods.
Learning dynamics with gradient methods can be analyzed considering learning rate $\to 0$ limit known as gradient flow~\cite{kushner2003stochastic}.
Therefore, we consider the following gradient flow
\begin{align}\label{eq:grad_flow}
    \frac{dz}{d\tau} = -\frac{\partial L}{\partial z},
\end{align}
using the loss function introduced above.
Here, $\tau$ denotes a time variable for learning dynamics, which should not be confused with $t$ representing the state transition.
Our aim is to investigate the convergence rate of a solution of the differential equation \req{eq:grad_flow} when $\sigma$ in the forget gate is replaced with another function $\phi$.

\subsection{Order of Saturation}\label{subsec:saturation_order}
To investigate the effect of choice of  gate functions on the convergence rate, we first define the candidate set $\mathcal F$ of bounded functions for the gate function.
\begin{definition}
    Let $\mathcal F$ be a set of differentiable and strictly increasing surjective functions $\phi:\mathbb R \to [0,1]$ such that 
    the derivative $\phi'$ 
    is monotone on $z > z_0$ for some $z_0 \ge 0$.
\end{definition}
$\mathcal F$ is a natural class of gating activation functions including 
$\sigma$.
As we explained in \rsec{subsec:saturation}, gated RNNs suffer from gradient vanishing due to saturation when learning long time scales.
To clarify the issue, we first show that saturation is inevitable regardless of the choice of $\phi \in \mathcal F$.
\begin{proposition}\label{prop:bounded_is_saturating}
  $\lim_{z\to \infty} \phi'(z)=0$ holds for any $\phi \in \mathcal F$.
\end{proposition}

Nevertheless, choices of $\phi$ significantly affect the efficiency of the training.
When the target $\lambda_*$ takes an extreme value near boundaries, the efficiency of training should depend on 
the asymptotic behavior of $\phi(z)$ for $z\gg 0$, that is, the rate at which $\phi(z)$ converges as $z \to \infty$.
We call the convergence rate of $\phi(z)$ as $z\to \infty$ as the \emph{order of saturation}.
More precisely, we define the notion as follows\footnote{Our definition for asymptotic order is slightly different from the usual one which adopts $\limsup_{z\to \infty} \frac{g(z)}{1-\phi(z)} < \infty$, since it is more suitable for analyzing training efficiency.}:
\begin{definition}
    Let $g:\mathbb R \to \mathbb R$ be a decreasing function. 
    $\phi \in \mathcal F$ has the \emph{order of saturation} of $O(g(z))$ if $\lim_{z\to \infty} \frac{g(az)}{1-\phi(z)} =0$ for some $a > 0$. 
    For $\phi, \tilde \phi \in \mathcal F$, $\phi$ has a \emph{higher order of saturation} than $\tilde \phi$ if
    $ 
    \lim_{z\to \infty} \frac{1-\phi(z)}{1-\tilde \phi(az)} = 0    
    $ 
    holds for any $a > 0$ and ${\tilde \phi}^{-1} (\phi(z))$ is convex for $z \gg 0$.
\end{definition}
Intuitively, the order of saturation of $O(g(z))$ means that the convergence rate of $\phi$ to 1 is bounded by the decay rate of $g$ up to constant multiplication of $z$. 
For example, the sigmoid function $\sigma$ satisfies $e^{-az}/(1-\sigma(z)) \to 0$ as $z\to \infty$ for any $a>1$,
thus has the exponential order of saturation $O(e^{-z})$.
The convexity condition for a higher order of saturation is rather technical, but automatically satisfied for typical functions, see Appendix~\ref{app:proof_theorem}.
If $\phi$ has a higher order of saturation (or saturates \emph{faster}) than another function $\tilde \phi$, then $\phi(z)$ converges faster than $\tilde \phi(z)$ as $z\to \infty$, and $\phi'(z)$ becomes smaller than $\tilde \phi'(z)$.
In this sense, training with $\tilde \phi$ seems more efficient than $\phi$ in the above problem.
However, this is not the case as we discuss in the next section.

\subsection{Efficient Learning via Fast Saturation}

To precisely analyze learning behavior, we trace the learning dynamics of the output value $f = \phi(z)$ 
since our purpose is to obtain the desired \emph{output} value rather than the input $z$.
We transform the learning dynamics (\req{eq:grad_flow}) into that of $f$ by
\begin{align}\label{eq:derive_dynamics_y}
    \frac{d f}{d\tau} = \frac{d z}{d \tau } \frac{d f}{d z}
    = - \phi'(z) \frac{\partial L}{\partial z}
    = - \phi'(z)^2 \frac{\partial L}{\partial f}.
\end{align}
To treat 
\req{eq:derive_dynamics_y} 
as purely of $f$, we define 
a function $g_\phi(f)$ of $f$ by
$ 
    g_\phi (f) := \phi'(\phi^{-1} (f)),
$ 
so that \req{eq:derive_dynamics_y} becomes
\begin{align}\label{eq:dynamics_y}
    \frac{d f}{d \tau} = - g_\phi(f)^2 \frac{\partial L}{\partial f}.
\end{align}
Our interest is in the dynamics of $f$ near the boundary, i.e., the limit of $f\to 1$.
We have the following result:

\begin{theorem}\label{thm:efficient_learning}
  Let $\phi, \tilde \phi \in \mathcal F$. 
  If $\phi$ has a higher order of saturation than $\tilde \phi$, then 
  $g_\phi (f)/ g_{\tilde \phi} (f) \to \infty$ as $f \to 1$.
\end{theorem}

\begin{figure}[t]
    \centering
    \includegraphics[width=0.33\paperwidth]{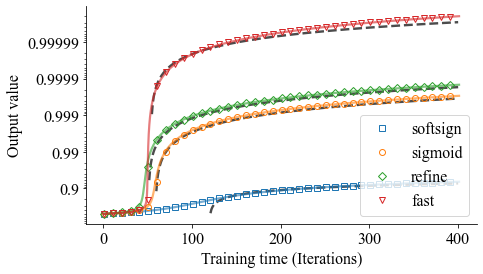}
    \caption{
    Learning curves for simplified long-time-scale learning problem with gradient descent (markers) and with gradient flow (solid lines).
    Gradient descent is done with learning rate 1.
    Time difference $t_1 - t_0$ is set to 10.
    Dashed lines are lower bounds given in \rtab{tab:gates_comparison}
    fitted to each learning curve with suitable translation.
    These lower bounds well approximate asymptotic convergence of gradient flow.
    Results of refine and fast gates are explained in \rsec{subsec:comparison}.
    }
    \label{fig:demo}
\end{figure}

Theorem~\ref{thm:efficient_learning} indicates that a higher order of saturation accelerates the move of the output $f$ near boundaries in accordance with \req{eq:dynamics_y} since $g_\phi(f)$ takes larger values.
Thus, contrarily to the intuition in \rsec{subsec:saturation_order}, 
a higher order of saturation leads to more efficient training for target values near boundaries.
We demonstrate this effect using two activation functions, the sigmoid function $\sigma(z)$ and normalized softsign function $\sigma_{\rm ns}(z) = (\mathrm{softsign}(z/2)+1)/2$ where $\mathrm{softsign}(z) = z / (1+|z|)$. 
$\sigma_{\rm ns}$ is the softsign function modified so that $0 \le \sigma_{\rm ns}(z) \le 1$ and $\sigma_{\rm ns}'(0) = \sigma'(0)$.
$\sigma$ has a higher order of saturation than $\sigma_{\rm ns}$ since $\sigma$ has the order of saturation of $O(e^{-z})$ and $\sigma_{\rm ns}$ has $O(z^{-1})$ (see \rfig{fig:saturation}).
We plot the learning dynamics of gradient flow for the problem in \rfig{fig:demo}.
Since $\sigma$ has a higher order of saturation than $\sigma_{\rm ns}$, 
the gate value $f$ of $\sigma_{\rm ns}$ converges slower to the boundary.
\rfig{fig:demo} also shows the dynamics of gradient descent with the learning rate 1.
While gradient descent is a discrete approximation of gradient flow, 
it behaves similar to gradient flow.

\textbf{Explicit convergence rates.}
Beyond Theorem~\ref{thm:efficient_learning},
we can explicitly calculate effective bounds of the convergence rate for the problem when the activation function is the sigmoid function $\sigma(z)$ or normalized softsign function $\sigma_{\rm ns}(z)$.

\begin{proposition}\label{prop:sigmoid_softsign_convergence_rate}
  Consider the problem in~\rsec{subsec:toy-problem} with the absolute loss $L = |f^{t_1-t_0} - \lambda_*|$ with $\lambda_* = 1$.
  For the sigmoid function $f = \sigma(z)$, the convergence rate for the problem is bounded as $1 - f = O(\tau^{-1})$. Similarly, for the normalized softsign function $f = \sigma_{\rm ns}(z)$, the convergence rate is bounded as $1 - f = O(\tau^{-1/3})$.
\end{proposition}

Proposition~\ref{prop:sigmoid_softsign_convergence_rate} shows the quantitative effect of difference in the order of saturation on the convergence rates.
We fit the bounds to the learning curves with the gradient flow in \rfig{fig:demo}.
The convergence rates of the learning are well approximated by the bounds.
These asymptotic analysis highlights that choices of the function $\phi$ significantly affects efficiency of training for long time scales.

\section{Proposed Method}\label{sec:proposed_method}

On the basis of the analysis in \rsec{sec:saturation_learnability}, we construct the fast gate, which is suitable for learning long time scales.

\subsection{Desirable Properties for Gate Functions}\label{subsec:properties}

We consider modification of the usual sigmoid function to another function $\phi \in \mathcal F$ for the forget gate in a gated RNN.
Function $\phi$ should satisfy the following conditions.
\begin{enumerate}[(i)]
    \item $\phi$ has a higher order of saturation than $\sigma$, 
    \item $\phi(z) \approx \sigma(z)$ for $z\approx 0$,
    \item $\phi$ is symmetric in a sense that $\phi(-z) = 1 - \phi(z)$. 
\end{enumerate}
Condition (i) comes from the argument in the previous section that fast saturating functions learn values near boundaries efficiently.
Conditions (ii) and (iii) indicate that the 
function $\phi(z)$ behaves similarly to $\sigma(z)$ around $z=0$.
In order to avoid possible harmful effects due to the modification, we do not want to change the behavior of the function away from the saturating regime.
Hence, we require these conditions.
The requirements are analogous to those by \citet[Section~3.4]{gua2020improving} for the gate adjustment.
The first condition can be viewed as a theoretical refinement of their heuristic modification.

\subsection{Fast Gate}

We explore gate functions satisfying the above conditions.
Recall that the sigmoid function $\sigma(z)$ has the exponential order of saturation.
From condition (i) in the previous section, we explore functions saturating superexponentially.
Since any superexponential order can be written as $O(e^{-\alpha(z)})$ with a function satisfying $\alpha(z)>z$ for large $z$,
it is enough to consider a function of the form
$ 
    \phi(z) = \sigma(\alpha(z))
$ 
for such $\alpha$.
The desirable properties in \rsec{subsec:properties} are rephrased as follows in terms of $\alpha$: (i) $\alpha(z) \gg z$ for $z \gg 0$, (ii) $\alpha'(0) = 1$, and (iii) $\alpha(-z)$ = $-\alpha(z)$ for $z \in \mathbb R$.
Such functions can be found as examples in the form
$ 
    \alpha(z) = z + p(z)
$ 
where $p$ is a polynomial consisting of only odd higher degree terms, such as $\alpha(z) = z + z^3$.
Since a higher degree term has a larger effect on the order of saturation, it mitigates gradient vanishing of the gate function more in accordance with Theorem~\ref{thm:efficient_learning}.
Thus, we take a limit of the degree to infinity, which leads to a simple function expression
\begin{align}
    \alpha(z) &= z + \frac{z^3}{3!} + \frac{z^5}{5!} + \cdots   \\
              &= \sinh(z) := \frac{e^z - e^{-z}}{2}.
\end{align}
Therefore, we adopt $\alpha(z) = \sinh(z)$ for the alternative function $\phi = \sigma \circ \alpha$ and obtain the fast gate
\begin{align}
    \phi (z) := \sigma (\sinh (z)).
\end{align}
This simple expression enables us to implement it with only one additional function.
Note that the above form is an example of gate functions which satisfy desirable properties in \rsec{subsec:properties} and there are infinitely many other possible choices.
The above particular form is one of the simplest choices, and not a limitation of our method.
For discussion of other candidates, see Appendix~\ref{app:choice_of_gate}.

\begin{table}[bt]
    \centering
    \resizebox{\linewidth}{!}{
    \begin{tabular}{ccccc}
        \hline
         & Softsign & Sigmoid & Refine & Fast \\
        \hline 
        \makecell{Saturation \\ order} & $O(z^{-1})$  & $O(e^{-z})$  & $O(e^{-2z})$ & $O(e^{-e^z})$ \\
        \hline 
        \makecell{Convergence \\ rate} & $O(\tau^{-\frac{1}{3}})$ & $O(\tau^{-1})$ & $O(\tau^{-1})$  & $O(W^2(c\tau^{-\frac{1}{2}}))$ \\
        \hline
        \makecell{Additional \\ parameters} & No & No & Yes & No \\
        \hline
    \end{tabular}
    }
    \caption{Comparison of order of saturation and convergence rate for problem in \rsec{sec:saturation_learnability}.
    ``Fast" denotes our method.
    $W^2(\cdot)$ is square of Lambert's function $W$ defined as inverse of function $z \mapsto ze^z$. $c>0$ is some constant.}
    \label{tab:gates_comparison}
\end{table}

\subsection{Comparison with Other Gate Functions}\label{subsec:comparison}

We analyze the fast gate $\phi(z)$ and compare it with other gate functions.
First, the order of saturation of the fast gate is $O(e^{-e^{z}})$ since
$e^{-e^{az}} / (1-\phi(z)) \to 0$ as $z\to \infty$ for any $a>1$.
We briefly describe the method of the refine gate~\cite{gua2020improving}, which was proposed to avoid gradient vanishing of the gate function.
This method exploits an auxiliary gate $r_t \in [0,1]^n$ to modify the forget gate value $f_t$ to $g_t = r_t( 1 - (1-f_t)^2) + (1-r_t) f_t^2$.
When a large output value is desired for the forget gate value, the auxiliary gate is expected to take $r_t \approx 1$ to push $f_t$ to $g_t \approx 1- (1-f_t)^2$.
Therefore, from the asymptotic view point, this method modifies the order of saturation of the gate function from $O(e^{-z})$ to $O(e^{-2z})$.
Compared with the refine gate, the fast gate has a much higher order of saturation.
We also analyze the asymptotic convergence rates of solving the toy problem in \rsec{sec:saturation_learnability}.
We summarize the results in \rtab{tab:gates_comparison}. 
See Appendix~\ref{app:convergence_rate} for detailed derivation.
Since the fast gate has a doubly exponential order of saturation, the order of convergence rate of learning long time scales is faster than the sigmoid and refine gates which have an exponential order of saturation 
(see also \rfig{fig:demo} for comparison of the convergence rates).
In addition, the fast gate does not require additional parameters whereas the refine gate does.
Therefore, the fast gate is computationally more efficient than the refine gate.

\section{Experiments}\label{sec:experiments}


\subsection{Synthetic Tasks}\label{subsec:synthetic_task}

\begin{figure}[t]
    \centering
    \includegraphics[width=1.\linewidth]{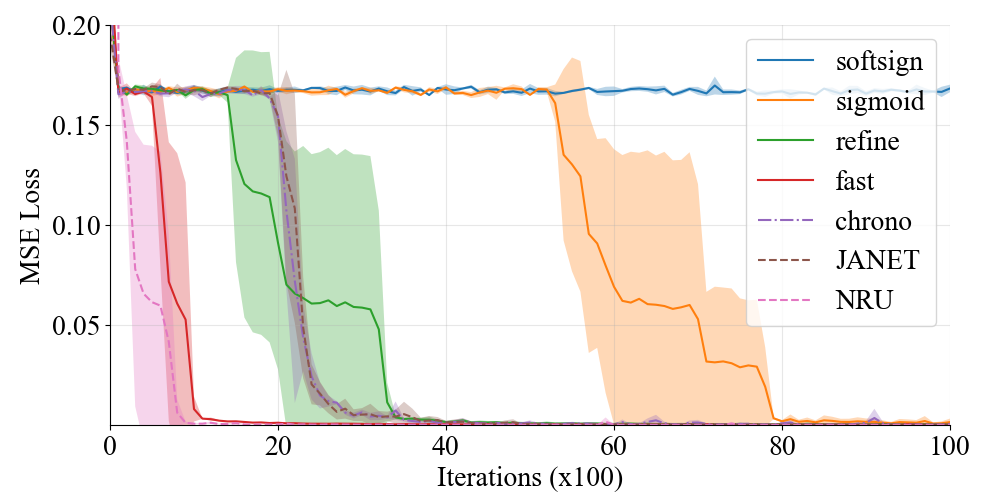}
    \caption{
        MSE loss for adding task of sequence length 5000
    }
    \label{fig:adding_mse}
\end{figure}

\begin{figure}[t]
    \centering
    \includegraphics[width=.98\linewidth]{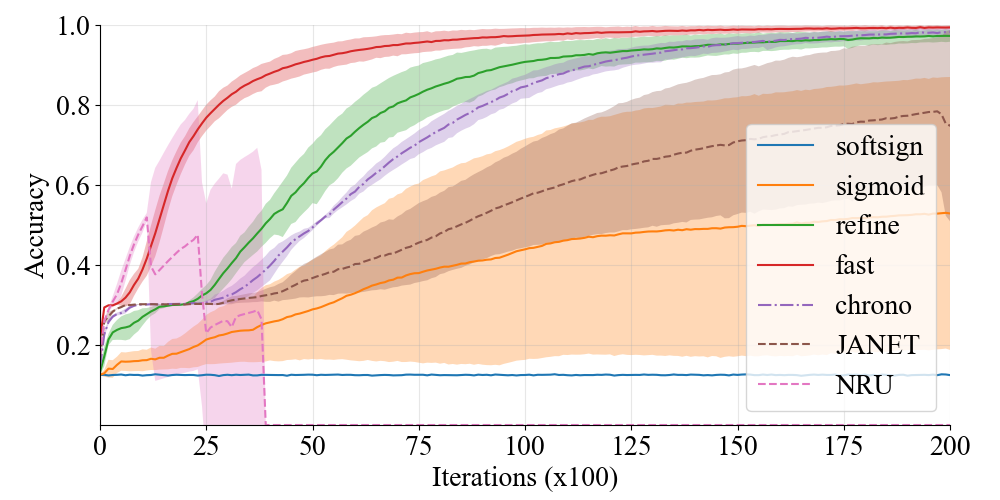}
    \caption{
        Accuracy for copy task of sequence length 500
    }
    \label{fig:copy_accuracy}
\end{figure}

\begin{figure}[t]
    \centering
    \includegraphics[width=.98\linewidth]{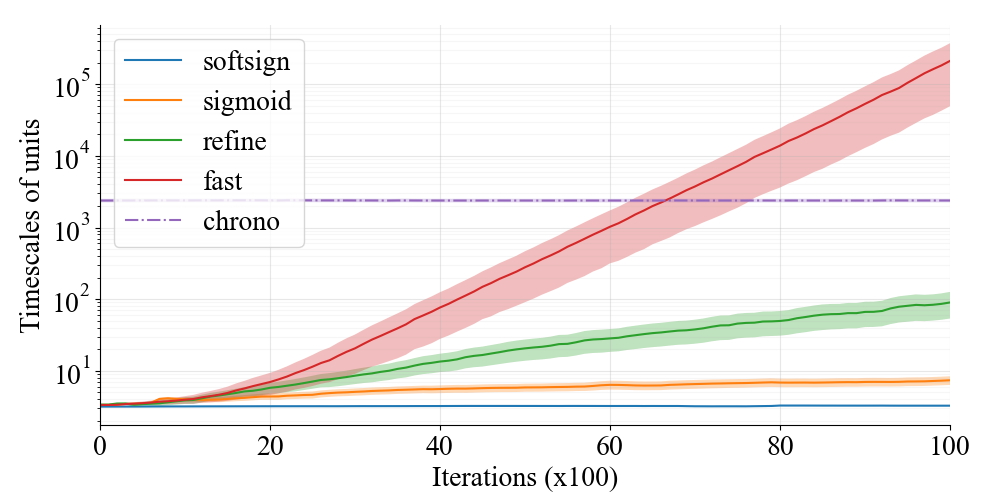}
    \caption{
        Growth of time scales of units over various gate functions on adding task.
        Lines and shaded areas represent mean and standard deviation of time scales over 128 units in state vector of models.
        Standard deviation is multiplied by 0.1 for visibility.
        Fast gate learns exponentially larger magnitudes of time scales than other gates.
    }
    \label{fig:adding_timescale}
\end{figure}

We evaluate the learnability for long time scales across various methods on two synthetic tasks, adding and copy, following previous studies~\cite{hochreiter1997long, arjovsky2016unitary}.
While these tasks are simple and easy to solve for short sequences,
they get extremely difficult for gated RNNs to solve when the sequence length grows to hundreds or thousands.

\textbf{Setup.}
We compare the fast gate with the refine gate because \citet{gua2020improving} reported that the refine gate achieved the best performance among other previous gate variants.
We also include the normalized softsign function $\sigma_{\rm ns}(z)$ in \rsec{sec:saturation_learnability} as a referential baseline to test the compatibility of our theory.
We use these gate functions in a single-layer LSTM.
Since the initialization of the forget gate bias $b_f$ is critical to model performance~\cite{tallec2018can},
we set it so that $\phi(b_f) = 1/(1+e^{-1})$ is satisfied for each gate function $\phi$ (with the bias for an additional gate function initialized by 0 for the refine gate), which amounts to $b_f =1$ in the usual sigmoid case~\cite{gers1999learning,greff2016lstm}.
We also compare performance of these gate variants to that of \textbf{chrono}-initialization~\cite{tallec2018can}, a method to initialize parameters to represent long time scales for the sigmoid gate. 
In addition to the LSTM, we include \textbf{JANET}~\cite{van2018unreasonable} and \textbf{NRU}~\cite{chandar2019towards} as baselines.
JANET is one of the simplest gated RNNs specialized to learn long time scales by omitting gates other than the forget gate and applying chrono-initialization. 
NRU uses non-saturating activation functions to write or erase to a memory cell.
We train and evaluate each model three times by varying the random seed. 
See Appendix~\ref{app:synthetic_task} for detailed setting. 

\textbf{Results.}
The mean squared error on the adding task of sequence length 5000 and the accuracy on the copy task of sequence length 500 during training are shown in \rfig{fig:adding_mse} and \ref{fig:copy_accuracy}, respectively.
While NRU requires the least number of parameter updates on the adding task, the training diverges on the copy tasks due to gradient explosion~\cite{pascanu2013difficulty}. 
This is because the state in the NRU evolves on an unbounded region; thus, a small parameter update can drastically change the behavior of the model.
We could not fix this instability even by reducing the clipping threshold for gradient by a factor of $10$.
We hypothesize that the training of the NRU tends to be more unstable on the copy task because this task has higher dimensional nature than the adding task in the sense of input dimension (10 vs 2) and the number of tokens to memorize (10 vs 2).
Among the gate functions, the fast gate converges the fastest.
This is due to the higher order of saturation:
the fast gate has the order of saturation $O(e^{-e^{z}})$ whereas the refine gate has $O(e^{-2z})$, thus learns long time scales more efficiently.
The normalized softsign gate completely fails to learn since it has a lower order of saturation $O(z^{-1})$ than the sigmoid function.
Thus, the performance of the models is well explained by the difference in the order of saturation in \rtab{tab:gates_comparison}.
This indicates that our theoretical analysis matches practical settings despite the fact that it builds on a simplified learning problem.
The result also shows that modification of the gate function is more effective for learning long-term dependencies than other methods such as chrono-initialization and JANET.

We further observe the growth of time-scale distribution of the memory cell in the LSTM on the adding task.
The time scale of $i$-th unit in the memory cell is measured using the bias term of the forget gate by $-1/\log \phi(b_{f,i})$~\cite{tallec2018can,mahto2020multi}.
We show the statistics of time scales over all 128 units at each iteration of the training in \rfig{fig:adding_timescale}.
The fast gate represents much longer time scales than other gate functions after training, which validates our method.
While chrono-initialization set time scales so that they are uniformly distributed within the range $[1, 5000]$, 
they do not change at all during training due to saturation of the usual sigmoid function $\sigma(z)$.
Since the fast gate can learn to adapt to even longer time scales than such initialization, it is effective to approximate arbitrary desired time scales which is usually unknown a priori.

\subsection{Pixel-by-pixel Image Recognition}

\begin{table}[t]
    \centering
    \resizebox{\linewidth}{!}{
    \begin{tabular}{ccccc}
        \hline
        & sMNIST & psMNIST & sCIFAR & Time \\
        \hline 
        Softsign      & 97.50 $\pm$ 0.58 & 91.71 $\pm$ 0.33 & 59.21 $\pm$ 0.39 & 17.7 min.\\
        Sigmoid       & 98.88 $\pm$ 0.12  & 95.71 $\pm$ 0.02 & 69.14 $\pm$ 0.39 & 14.3 min.\\
        Refine        & 98.94 $\pm$ 0.03  & 95.93 $\pm$ 0.16 & 69.55 $\pm$ 0.50 & 22.7 min.\\
        \textbf{Fast} & 99.05 $\pm$ 0.04  & 96.18 $\pm$ 0.14 & 70.06 $\pm$ 0.38 & 14.7 min.\\
        \hline
        chrono        & 98.83 $\pm$ 0.09  & 94.37 $\pm$ 0.69 & 60.36 $\pm$ 0.51 & 14.3 min. \\
        JANET         & 98.59 $\pm$ 0.03  & 93.85 $\pm$ 0.23 & 60.99 $\pm$ 0.51 & 10.6 min.\\
        NRU           & 98.73 $\pm$ 0.27 & 94.76 $\pm$ 0.35 & 62.32 $\pm$ 0.30 & 35.0 min.\\
        \hline
    \end{tabular}
    }
    \caption{Test accuracy on image classification tasks and processing time per epoch on psMNIST
    }
    \label{tab:sMNIST_results}
\end{table}

\begin{figure}[t]
    \centering
    \includegraphics[width=1.0\linewidth]{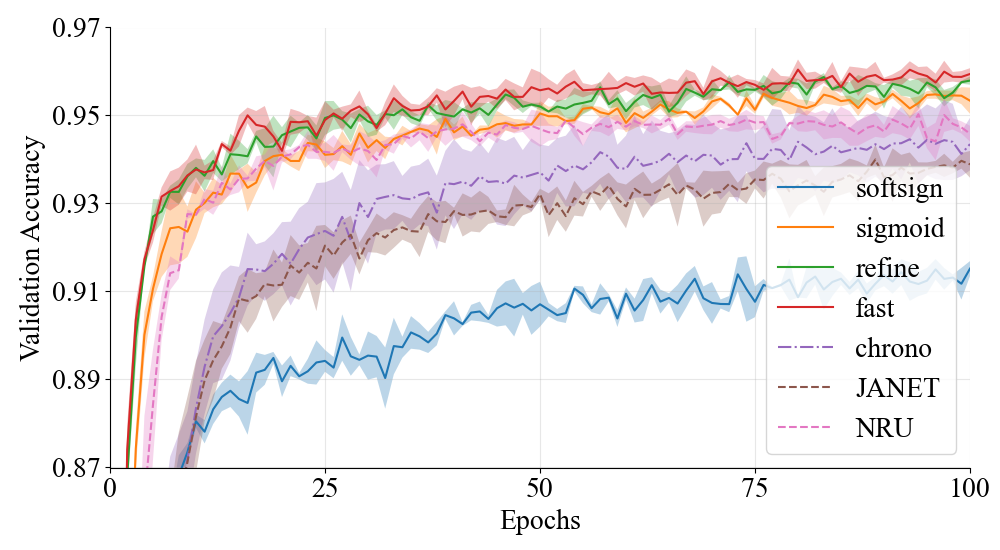}
    \caption{
        Validation accuracy for psMNIST
    }
    \label{fig:psMNIST_result}
\end{figure}

Next, we evaluate the fast gate on the sequential image recognition task,
where a pixel value is applied into recurrent models at each time step~\cite{le2015simple}.

\textbf{Setup.}
We use the usual order sequential MNIST (sMNIST) task and permuted order version (psMNIST) to introduce more complex and long-term temporal dependencies.
We also use the sequential CIFAR-10 (sCIFAR) task, which involves higher dimensional inputs and longer sequence length (i.e., $3 \times 1024$) than MNIST.
We train the LSTM with the various gates, JANET, and NRU.
See Appendix~\ref{app:image_classification} for training details.
We set the hidden dimension as 512 on all models and the dimension of the memory cell in NRU as 144 following the original paper.
We evaluate the averaged test accuracy with the standard deviation over three random seeds.
We also report the computational time to train each model to compare the computational efficiency of the LSTM with the fast gate to other models.

\textbf{Results.}
The results are shown in \rtab{tab:sMNIST_results}.
The fast gate performs the best among various gates while the normalized softsign gate poorly performs.
This is because the order of saturation of activation functions in the forget gate directly affects the learnability for long time scales.
The LSTM with the fast gate also outperforms all the other baselines.
Note that as chrono-initialization gives too heavy-tailed time scale distribution~\cite{gua2020improving}, the chrono-LSTM and JANET performs worse than simply initializing the gate bias as $b_f=1$ (Sigmoid in the table).
Since large model size tends to result in high accuracy on these tasks~\cite{voelker2019legendre,erichson2020lipschitz},
we also provide results of smaller models in Appendix~\ref{app:image_experiment_small} for comparison.
While the NRU achieves the highest accuracies on the psMNIST task in \rtab{tab:sMNIST_small_results} in the appendix, 
the NRU performs worse than the LSTMs in \rtab{tab:sMNIST_results}.
Therefore, performance of the LSTM seems to scale better than the NRU in model size. 
The refine gate requires more than 1.5 times longer processing time than the fast gate for training since it involves an auxiliary gate computation with additional parameters. 
The NRU requires longer processing time than the LSTMs due to its complicated state update rule.
The learning curve for the psMNIST task in \rfig{fig:psMNIST_result} shows that 
the fast gate reaches high accuracy in as few epochs as the refine gate.
Combining with the results on the processing time per epoch, we conclude that the fast gate learns complex time scales the most efficiently among the gate functions.

\begin{table}[t!]
    \centering
    \resizebox{\linewidth}{!}{
    \begin{tabular}{ccccccc}
        \hline
         & $>$ 10K & 1K-10K & 100-1K & $<$ 100 & All tokens & Time \\
        \hline
        Sigmoid & 7.25 & 27.72 & 170.25 & 2026.87 & 60.22       & 130 sec. \\ 
        Refine  & 7.61 & 28.01 & 166.46 & 1936.02 & 60.50       & 185 sec. \\
        \textbf{Fast} & 7.43 & 27.70 & 166.68 & 1975.51 & 60.09 & 138 sec.  \\
        \hline
    \end{tabular}
    }
    \caption{Test perplexity for tokens across different frequency bins on Penn Treebank dataset with training time per epoch 
    }
    \label{tab:language_results}
\end{table}

\subsection{Language Modeling}

In natural language processing, 
performance of language models can suffer from difficulty in learning long time scales because predicting statistically rare words involves long time scales~\cite{mahto2020multi}.
It is expected that using a forget gate function with a higher order of saturation improves learning to predict such rare words.
We validate this effect 
in the following experiment.

\textbf{Setup.}
We train and evaluate three-layer LSTM language models following a previous study \cite{mahto2020multi}\footnote{
https://github.com/HuthLab/multi-timescale-LSTM-LMs.}
on the Penn Treebank (PTB) dataset,
replacing every sigmoid forget gate in the baseline LSTM with the fast gate.
We compare the LSTM with the fast gate against the LSTM with the sigmoid and refine gates and also NRU by replacing all LSTM modules with NRU modules under the same experimental setting.
To evaluate the model performance on data involving different ranges of time scales, the test dataset is divided into four bins depending on their frequencies in the training dataset: more than 10,000, 1000-10,000, 100-1000, and fewer
than 100 occurrences.

\textbf{Results.}
The results are shown in \rtab{tab:language_results}.
The result for the NRU is not in the table since the training diverges. 
Both refine and fast gates improve perplexity for less frequently observed words compared to the sigmoid gate.
The model with the fast gate also achieves the lowest total perplexity.
Since frequently observed words involve short-term dependencies, this result indicates that the fast gate improves model performance by learning a wide range of time scales that appear in practical tasks. 
\rtab{tab:language_results} also shows the training time taken for one epoch.
We observe that the refine gate has larger computational overhead than the fast gate,
although the LSTM with the refine gate has the same number of parameters as other LSTMs by using the gate-tying trick  (Appendix~\ref{app:gate_tying}).
This is due to the two-stage computation for the gate value via the auxiliary gate, which cannot be parallelized, thus leads to slow computation.
In summary, the fast gate can improve performance on data involving extremely long time scales without sacrificing the performance on data involving short time scales and with less computational overhead.

\section{Conclusion}\label{sec:conclusion}

We analyzed the saturation problem in learning of gate functions in recurrent models.
Against the common intuition that saturation of the activation function degrades training, we proved that strengthening the saturating behavior 
is effective in mitigating gradient vanishing of gate functions.
We proposed the fast gate, which has a doubly exponential order of convergence with respect to inputs, by simply composing the hyperbolic sinusoidal function to the usual sigmoid function.
We evaluated the trainability of the fast gate on data involving extremely long time scales.
We empirically showed that the fast gate improves accuracy on benchmark tasks with little computational overhead.
Our analytical approach is applicable to any other bounded activation functions that appear in the core of modules such as an attention mechanism.
Thus, we expect that it can improve learnability of other neural networks beyond recurrent models.

\bibliography{ref}

\clearpage

\appendix

\section{Further discussion of related work}\label{app:related_work}

There are plenty of research approaching to learning of long-term dependencies of time series data with RNNs.
We discuss relations between our work and other previous studies.

Several methods on properly constraining a weight matrix of a simple RNN without gating mechanism are proposed to alleviate the gradient vanishing problem~\cite{arjovsky2016unitary,zhang2018stabilizing,lezcano2019cheap,helfrich2020eigenvalue}.
These methods limits expressivity of models due to parameter constraint~\cite{vorontsov2017orthogonality} and thus are not usually used for practical tasks.
Our method takes a different approach for learning long-term dependencies through time scales of models and does not limit the model expressivity.

There is another approach to improve learnability of gated RNNs by reducing redundant parameters~\cite{collins2016capacity,chen2018dynamical,kusupati2019fastgrnn}.
Since the forget gate is the most critical component of gated RNNs, these \emph{minimal} RNNs are also equipped with the forget gate.
Therefore, our method is applicable to those minimal models to further enhance model performances.

Recently, continuous-time models based on differential equations and their discrete counterparts have been attracting attention to deal with extremely long (and possibly irregularly sampled) time series data~\cite{chang2019antisymmetricrnn,voelker2019legendre,gu2020hippo,lechner2020learning,gu2021combining,rusch2021unicornn}.
Some of these models utilize the sigmoid function to represent variable time steps for discretization.
Such use of the sigmoid function essentially has the same effect of the forget gate in LSTM and GRU in terms of time scales of models~\cite{tallec2018can,gu2020hippo}. 
Thus, our method replacing gate function can be applied to the sigmoid function representing time steps to further enhance the learnability for wide range of time scales.

\section{Relation to Gradient Vanishing due to Recurrence}
\label{app:relation_to_gradient_vanishing}

When learning long-term dependencies of time series data, 
recurrent models often suffer from gradient vanishing due to recurrent calculation~\cite{bengio1994learning,pascanu2013difficulty}.
We describe the relation between this gradient vanishing and the saturation problem in this section.
First, consider a general recurrent model with state update rule of form $h_t = G(x_t, h_{t-1}; \theta)$, where $h_t$ and $x_t$ are the state and input at time $t$ and $\theta$ denotes the parameter vector of the model.
During training, the parameter $\theta$ is updated using the gradient
\begin{align}
    \frac{\partial L}{\partial \theta} = \sum_t \frac{\partial h_t}{\partial \theta} \frac{\partial L}{\partial h_t},
\end{align}
where the right hand side is calculated by back-propagation through time (BPTT)
\begin{align}
    \frac{\partial L}{\partial h_t} = \frac{\partial h_{t+1}}{\partial h_t}\frac{\partial L}{\partial h_{t+1}}.
\end{align}
If the matrix $\frac{\partial h_{t+1}}{\partial h_t}$ is contractive, the gradient $\frac{\partial L}{\partial h_t}$ exponentially decays through this backward recurrence.
Then, the gradient for $\theta$ does not include information on data at time $t$ distant from the last time step, which leads to difficulty in learning long-term dependencies of data.

We now consider a gated model with the state update rule $h_t = f_t \odot h_{t-1} + i_t \odot F(x_t, h_{t-1})$ where $f_t$ and $i_t$ are forget and input gates and $F$ is some state update function.
Then, we have
\begin{align}
    \frac{\partial h_{t+1}}{\partial h_t} = \mathrm{diag} (f_t) + \cdots.
\end{align}
Assuming that the first term $\mathrm{diag} (f_t)$ has relatively large effect than other terms, the decay rate of the gradient $\frac{\partial L}{\partial h_t}$ is largely dominated by the distance of $f_t$ to 1.
This corresponds to the time scales of the gated model.
Namely, if the forget gate represents only short time scales, then the gradient vanishing due to recurrence tends to occur.
On the other hand, if the forget gate learns long time scales, then it mitigates the gradient vanishing and helps the model to learn to extract important features at distant time steps.
Thus, learnability of gated models for long time scales is critical to capture long term dependencies.
In this work, we focus on learnability for long time scales by considering the gradient vanishing due to saturation of bounded functions.
The distinction of two gradient vanishing problems is easy to see through \rfig{fig:demo}:
every gate learns slowly at early stage (until 40 iteration) because of gradient vanishing due to recurrence. After it gets through the plateau, it now faces the gradient vanishing due to saturation to learn long time scales.
In this way, the gradient vanishing due to recurrence and the gradient vanishing of bounded function relates closely to each other.

\section{Details on Theoretical Results}
\label{app:proofs}

\subsection{Proof of Proposition \ref{prop:bounded_is_saturating}}

\begin{proof}

Let $\phi$ be an element of $\mathcal F$.
Since $\phi'$ is monotone for $z\gg 0$, $\lim_{z\to \infty} \phi'(z)$ exists.
Since $\phi$ is increasing, we obtain $\lim_{z\to \infty} \phi'(z) \ge 0$.
If $a := \lim_{z\to \infty} \phi'(z) > 0$, then $\phi(z) = \phi(0) + \int_0^z \phi'(z) dz \ge \phi(0) + az$.
This contradicts the boundedness of $\phi$.
Thus, we have $\lim_{z\to \infty} \phi'(z) = 0$.
\end{proof}

\subsection{Proof of Theorem \ref{thm:efficient_learning}}\label{app:proof_theorem}

First, we remark some technical points on the following definition of the order of saturation.
\begin{definition}
    For $\phi, \tilde \phi \in \mathcal F$, $\phi$ has \emph{higher order of saturation} than $\tilde \phi$ if ${\tilde \phi}^{-1} (\phi(z))$ is convex for $z \gg 0$ and
    \begin{align}\label{eq:higher_order_saturation}
    \lim_{z\to \infty} \frac{1-\phi(z)}{1-\tilde \phi(az)} = 0    
    \end{align}
    holds for any $a > 0$.
\end{definition}
The first condition for convexity is automatically satisfied for typical functions such as logarithmic, polynomial, and exponential and required to ensure the growth rates of functions to be well behaved\footnote{We have not found any examples of $\phi, \tilde \phi \in \mathcal F$ which do not satisfy the convexity of ${\tilde \phi}^{-1} (\phi(z))$ under \req{eq:higher_order_saturation}.}.
For example, consider the fast gate $\phi(z) = 1/(1+e^{-\sinh (z)})$ and sigmoid function $\tilde \phi(z) = \sigma(z) = 1/(1+e^{-z})$.
Then ${\tilde \phi}^{-1} (\phi (z)) = \sinh (z)$ is indeed convex for $z>0$.
The second condition is slightly different from the common definition of order that adopts $\lim_{z\to \infty} \frac{\phi(z)}{\tilde \phi(z)} = 0$.
While the two limit conditions coincides for logarithmic and polynomial classes, there is a difference in the treatment of exponential functions.
For example, $O(e^z)$ and $O(e^{2z})$ are distinguished in the usual sense but are identified in our definition.
Note that in terms of learning of neural networks, this amounts to ignoring re-parametrization of parameters with constant multiplication $ \theta \to 2 \theta $.
Since such naive re-parametrization should not drastically change the learning behavior, our definition for the order of saturation makes more sense for considering learning of neural networks.
This is well illustrated in \rtab{tab:gates_comparison} and \rfig{fig:demo}:
the refine gate has the saturation of order $O(e^{-2z})$ but results in the same order of convergence rate of learning $O(\tau^{-1})$ as the usual sigmoid gate since $O(e^{-2z})$ is the same as $O(e^{-z})$ in our definition.
The fast gate, however, has a higher order of saturation in the sense of our definition, thus has a faster order of convergence rate.


We now prove Theorem~\ref{thm:efficient_learning}.

\begin{proof}

Take $\phi, \tilde \phi \in \mathcal F$ such that $\phi$ has a higher order of saturation
than $\tilde \phi$. 
Define a function $\eta$ by $\eta(z) := {\tilde \phi}^{-1} (\phi (z))$.
Note that $\eta(z)$ is convex for $z\gg 0$ since $\phi$ has a higher order of saturation than $\tilde \phi$.
Let $a > 0$ be any positive number.
Then, we have $\eta (z) > az$ for $z\gg 0$ since $\phi (z) > \tilde \phi(az)$ for $z \gg 0$ and $\tilde \phi$ is increasing.
Thus, for any large $z_0>0$, we get $\eta'(z) \ge a$ for some $z>z_0$.
By convexity of $\eta$ on $z \gg 0$, we obtain $\eta'(z) \ge a$ for $z \gg 0$.
On the other hand, we have
\begin{align}
    \eta'(z) &= \frac{ \phi'(z)}{\tilde \phi'(\tilde \phi^{-1}(\phi(z)))} \\
    &= \frac{g_\phi(\phi(z))}{g_{\tilde \phi}(\phi(z))} 
\end{align}
By substituting $f := \phi(z)$,
we obtain
\begin{align}
    \eta'(z) = \frac{g_\phi(f)}{g_{\tilde \phi} (f)}
    \ge a
\end{align}
for $f \approx 1$.
Since $a > 0$ is arbitrary, we have 
$\frac{g_\phi(f)}{g_{\tilde \phi}(f)} \to \infty$ as $f \to 1$.
\end{proof}

\subsection{Derivation of Explicit Convergence Rate}\label{app:convergence_rate}

In this section, we derive the convergence rates of learning long time scales on the toy problem in \rsec{sec:saturation_learnability} for the sigmoid and normalized softsign gate, proving Proposition~\ref{prop:sigmoid_softsign_convergence_rate}.
Similarly, we also derive the convergence rates for the refine and fast gate.
We consider the absolute loss $L = |\lambda_* - f^{t_1-t_0}|$ with the gate value $f=\phi(z)$.
Since the learning dynamics does not depend on $\lambda_*$ when we consider the absolute loss, we may take $\lambda_* = 1$ without loss of generality.

\textbf{The sigmoid gate case.}
We first analyze the case with the sigmoid gate $f=\sigma(z)$.
We assume that the initial value of $f$ is small so that $f^{t_1-t_0} < \lambda_*$.
Then \req{eq:dynamics_y} becomes
\begin{align}
    \frac{d f}{d \tau} = - \sigma'(z)^2 \frac{d L}{d f} = (t_1 - t_0) f^{t_1 - t_0 +1} (1-f)^2,
\end{align}
as long as $f^{t_1-t_0} < \lambda_*$ holds.
Since $f$ monotonically increases over time, we obtain a lower bound for the right hand side after time $\tau > \tau_0$ as $(t_1 - t_0) f^{t_1 - t_0 +1} (1-f)^2 > C (1-f)^2$ with 
$C=(t_1 - t_0) f(\tau_0)^{t_1 - t_0 +1}$.
This gives a lower bound of $f$ the dynamics of which is defined by
\begin{align}
    \frac{d f_{\rm lower}}{d \tau} = C (1-f_{\rm lower})^2.
\end{align}
Solving this, we obtain $1-f_{\rm lower} = 1 / C \tau$.
Thus, we obtain the upper bound of the difference
\begin{align}\label{eq:approx_sigmoid}
    1 - f = O(\tau^{-1}),
\end{align}
which approximates the asymptotic convergence rate for training.

\textbf{Tightness of bounds.}
In the above, we only evaluated the upper bound of the difference $1-f$ as the convergence rate.
We can easily show that this evaluation is tight and thus the bound gives the exact order of convergence rate.
We consider the sigmoid case for example.
While the lower bound of gradient is evaluated as $\frac{df}{d\tau} \ge C(1-f)^2$, it is also upper bounded by  $\frac{df}{d\tau} \le \tilde C(1-f)^2$ with another $\tilde C > 0$ since $(t_1 - t_0) f^{t_1-t_0+1}$ is upper bounded.
Thus, we have the same order of upper bound of $f$ and thus the bound gives tight order of convergence rate.
This argument easily holds in other cases below, but we omit it for simplicity.

\textbf{The normalized softsign gate case.}
For the asymptotic analysis, we may assume $z>0$.
Then the normalized softsign function is $\phi(z) = \sigma_{\rm ns}(z) = \frac{1}{2}(\frac{z}{|z|+2} + 1) = 1 - \frac{1}{z+2}$.
Thus, $\phi'(z) = \frac{1}{(z+2)^2} = (1 - \phi(z))^2$.
Then for $f = \phi(z)$, \req{eq:dynamics_y} becomes
\begin{align}
    \frac{df}{d\tau} = -\phi'(z)^2 \frac{dL}{df} = (t_1 - t_0) f^{t_1-t_0} (1-f)^4.
\end{align}
as long as $f^{t_1-t_0} < \lambda_*$ holds.
The lower bound of the right hand side over $\tau > \tau_0$ is given by $(t_1 - t_0) f^{t_1- t_0} (1-f)^4 > C(1-f)^4$ with constant $C= (t_1 - t_0) f(\tau_0)^{t_1- t_0}$ since $f$ increases during learning.
Therefore, the dynamics of a lower bound of $f$ can be written as
\begin{align}
    \frac{df_{\rm lower}}{d\tau} = C(1 - f_{\rm lower})^4.
\end{align}
Solving this, we get $1 - f_{\rm lower} = (C\tau)^{-1/3}$.
Thus, we obtain the upper bound of the difference
\begin{align}
    1 - f = O(\tau^{-1/3}).
\end{align}

\textbf{The refine gate case.}
The refine gate~\cite{gua2020improving} is an auxiliary gate $r = \sigma(\tilde z)$ that adjust the forget gate value $f=\sigma(z)$ to the effective gate value $g=2rf + (1-2r)f^2$.
We first reformulate the problem as minimizing the loss function
\begin{align}
    L = | \lambda_* - g^{t_1 - t_0}|,
\end{align}
i.e., we replace $f$ with $g$.
Then, 
the learning dynamics of $g$ is given as
\begin{align}
    \frac{dg}{d\tau} 
    &= \frac{d z}{d \tau} \frac{\partial g}{\partial z} + \frac{d \tilde z}{d \tau} \frac{\partial g}{\partial \tilde z} \\
    &= -\Bigl( (\frac{\partial g}{\partial z})^2 + (\frac{\partial g}{\partial \tilde z})^2 \Bigr) \frac{\partial L}{\partial g}. 
\end{align}
Since the refine gate $r$ monotonically increases during learning, we may assume $r\ge 1/2$.
We claim that $\frac{\partial g}{\partial z}$ and $\frac{\partial g}{\partial \tilde z}$ are bounded as follows when $r \ge 1/2$:
\begin{align}
    f(1-g) \le \frac{\partial g}{\partial z} \le 2f(1-g) \\
    0 \le \frac{\partial g}{\partial \tilde z} \le r (1-g)
\end{align}
Indeed, we have
\begin{align}
    \frac{\partial g}{\partial z}
    &= 2f(1-f)(r+(1-2r)f) \\
    &= 2f \bigl( 1-g - (1-r)(1-f) \bigl) 
    \ \Bigl( \le 2f(1-g) \Bigr) \\
    &= f \bigl( 1 - g + (2r-1)(1-f)\bigr) 
    \ \Bigl( \ge f(1-g) \Bigr)
\end{align}
and
\begin{align}
    \frac{\partial g}{\partial \tilde z}
    &= 2fr(1-r) - 2f^2 r (1-r)  \\
    &= r \bigl( 1-g - (1-f)^2 \bigr) \le r(1-g).
\end{align}
Thus, we get a bound
\begin{align}
    C(1-g)^2 \le \frac{dg}{d\tau} 
    \le \tilde C (1-g)^2
\end{align}
with $C = (t_1-t_0)g(\tau_0)^{t_1-t_0-1} f(\tau_0)^2$ and $\tilde C = 5 (t_1-t_0)g(\tau_0)^{t_1-t_0-1}$.
With this bound, we can do the exactly same calculation as the sigmoid case to obtain the bound
\begin{align}
    1-g = O(\tau^{-1}).
\end{align}

\textbf{The fast gate case.}
Let $\phi(z) = \sigma(\sinh(z))$ be the fast gate.
By the chain rule, the derivative of the function $\phi$ is given by
$\phi'(z) = \phi(z) (1 - \phi(z)) \cosh(z)$.
Therefore, \req{eq:dynamics_y} for $f = \phi(z)$ can be written as
\begin{align}
    \frac{df}{d\tau} 
    &= -(\phi'(z))^2 \frac{\partial L}{\partial f} \\
    &= f^2 (1-f)^2 \cosh^2(z) (t_1 - t_0) f^{t_1 - t_0 -1} \\
    &= (t_1 - t_0) f^{t_1 - t_0 + 1} (1 - f)^2 (1 + \sinh^2(z)) \\
    &\ge C(1-f)^2 \sinh^2(z)
\end{align}
with constant $C := (t_1-t_0) f(\tau_0)^{t_1 - t_0 + 1}$.
Since we consider the asymptotic convergence rate, we assume $f$ is sufficiently close to 1 so that $\sinh(z) = \sigma^{-1}(f) = \log (\frac{1}{1-f} -1) \ge -\frac{1}{2}\log(1-f)$ 
and $\frac{\log (1-f)}{2+\log (1-f)} \le 2$.
Then, by properly replacing the constant $C$, we obtain
\begin{align}
    \frac{df}{d\tau} 
    &\ge C (1-f)^2 (\log(1-f))^2 \frac{\log(1-f)}{2 + \log(1-f)}.
\end{align}
This inequality gives the upper bound $y$ of the difference $1-f$, the dynamics of which is given by
\begin{align}
    \frac{dy}{d\tau} = -C y^2 (\log y)^2 \frac{\log y}{2 + \log y}.
\end{align}
Rewriting this equation, we have
\begin{align}
    \frac{2 + \log y}{y^2 (\log y)^3}\frac{dy}{d\tau} = C.
\end{align}
By integration, we get
\begin{align}
    \frac{1}{y (\log y)^2} = C\tau.
\end{align}
Thus we obtain
\begin{align}
    y^{1/2} \log y^{1/2} = (4C\tau)^{-1/2},
\end{align}
or equivalently,
\begin{align}
    y = W((4C\tau)^{-1/2})^2,
\end{align}
where $W$ is an inverse of function $z \mapsto z \log z$.
This gives the effective bound of the difference $1-f$ as
\begin{align}
    1-f = O(W(c\tau^{-1/2})^2)
\end{align}
with some constant $c$.
Note that this order is asymptotically smaller than $O(\tau^{-1})$ which is the convergence rate of learning for the sigmoid and refine gate.

\subsection{Further Visualization}

\begin{figure}[t]
    \centering
    \includegraphics[width=0.8\linewidth]{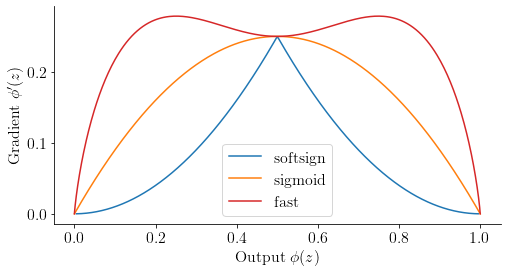}
    \caption{Gradient of gate functions with respect to output value, which is equivalent to $g_\phi$ in \rsec{subsec:saturation_order}. Functions with higher order of saturation have larger gradient at both ends.}
    \label{fig:gates_gradient}
\end{figure}

\rfig{fig:gates_gradient} shows the increase in the gradient of gate functions with different order of saturation against output values.
This illustrates the intuition for Theorem~\ref{thm:efficient_learning} that the gradient $g_\phi (f) = \phi'(\phi^{-1}(f))$ with respect to the output $f$ has different convergence rate to the limit $f\to 1$ in accordance with the order of saturation.

\begin{figure}[t]
  \begin{minipage}[b]{0.49\linewidth}
    \centering
    \includegraphics[width=\linewidth]{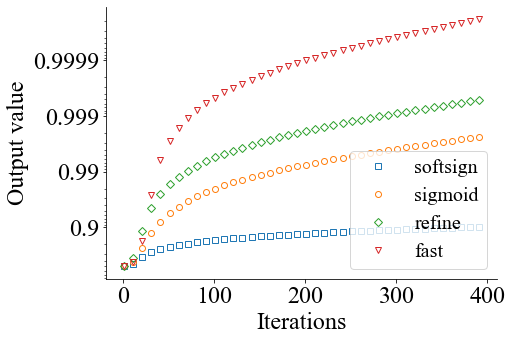}
    \subcaption{RMSprop}
    \label{fig:demo_rmsprop}
  \end{minipage}
  \begin{minipage}[b]{0.48\linewidth}
    \centering
    \includegraphics[width=\linewidth]{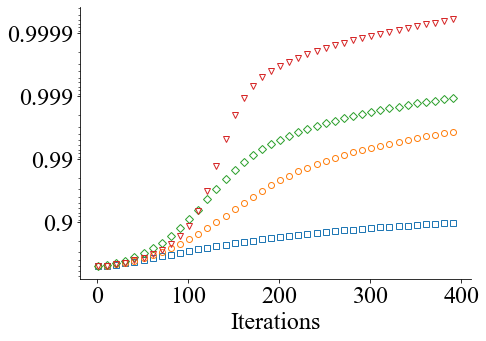}
    \subcaption{Adam}
    \label{fig:demo_adam}
  \end{minipage}
  \caption{Learning dynamics for toy problem in \rsec{sec:saturation_learnability} with RMSprop and Adam. The fast gate consistently learns the fastest.}
  \label{fig:demo_other}
\end{figure}

In addition to the simple gradient descent, we also consider solving the toy problem in \rsec{sec:saturation_learnability} with other optimizers.
RMSprop and Adam optimizer~\cite{kingma2014adam} are widely used for training deep neural networks.
We plot the learning dynamics on the toy problem with RMSprop and Adam in \rfig{fig:demo_other}.
Since RMSprop and Adam solve the problem more efficiently than the gradient descent, we set the learning rate $0.01$ and $t_1-t_0 = 30$.
Other hyperparameters are set to the default in pytorch library.
We observe that the fast gate consistently learns faster than other gate functions.
Thus, we conclude that our method also applies to training of gated recurrent networks with practical optimizers.

\section{Arbitrariness of Fast Gate}\label{app:choice_of_gate}

In \rsec{sec:proposed_method}, we adopted the particular form $\phi(z) = \sigma(\sinh z)$ as the fast gate.
There are infinitely many other choices of gate functions satisfying the desirable properties in \rsec{subsec:properties}.
For example, we can construct a function saturating even faster by composing $sinh$ again, i.e., $\phi(z) = \sigma(\sinh(\sinh(z)))$.
A natural question is whether such a \emph{faster} gate function further improves the performance of gated models.
We evaluated the LSTM with the iterated fast gate function (\textbf{iter\_fast}) on the synthetic and image classification tasks.
The training converges even faster on both adding and copy tasks~(\rfig{fig:adding_runtime}, \ref{fig:copy_acc_runtime}), which is consistent with our theory.
On the other hand, there are almost no difference in test accuracy of the image classification tasks~(``Iterated Fast LSTM'' in \rtab{tab:sMNIST_results_GRU}), possibly because the models must learn other complex features in addition to time scales.
Note that the processing time increases by the application of another $\sinh$ function.
Taking the results and the overhead of additional $\sinh$ into account, only one composition of $\sinh$ (i.e., the fast gate in the main text) seems practically the simplest and most efficient solution for learning long time scales.
Further investigation of other effective candidates is left as future work.

\section{Experimental Details}\label{app:experiment}

\subsection{Computational settings}
Our computational setup is as follows: CPU is Intel Xeon Silver 4214R 2.40GHz, memory size is 512 GB, and the GPU is NVIDIA Tesla V100S.

\subsection{LSTM with Gate Tying and Parameter Count}\label{app:gate_tying}

An LSTM has four affine transform computation in its state update rule as in \req{eq:lstm}.
Since each affine transform has the same number of parameters, the LSTM has $4N$ parameters, where $N$ is the number of parameters for one linear transform.
With a thorough exploration on variants of gated recurrent models, \citet{greff2016lstm} reported that coupling the forget gate with the input gate by 
\begin{align}
    f_t = 1_n - i_t
\end{align}
does not empirically degrade model performance.
Here $n$ is the hidden dimension and $1_n \in \mathbb R^n$ is an $n$-dimensional vector which has 1 for all entries.
This trick reduces the model parameters from $4N$ to $3N$.
\citet{gua2020improving} proposed the refine gate on the basis of this variant of LSTM.
Their method introduces additional $N$ parameters to train an auxiliary gate; thus the total number of model parameters matches to the $4N$ of the original LSTM.
In the experiments excluding language modeling,
we used the gate-tied variant of LSTM to fairly compare the learnability of the gate functions. 
Thus, the LSTMs with the normalized softsign, sigmoid, and fast gates have $3N$ parameters while that with the refine gate LSTM has $4N$.
In the language-modeling experiment, we applied the fast gate to the forget gate without tying gates to compare the practical performance on an equal number of parameters.
Note that the JANET has $2N$ parameters because it further removes the output gate from the gate-tied LSTM.

\subsection{Synthetic Tasks}\label{app:synthetic_task}

The adding and copy tasks are defined as follows.

\textbf{Adding task.}
The input for this task consist of two sequences of prescribed length $T$. One is a sequence of random numbers $(a_1, \cdots, a_T) \in \mathbb R^T$ sampled from the uniform distribution on $[0,1]$. The other is the \emph{indicator} sequence, which has 1 on two random indices $i_1 \in [1,T/2)$ and $i_2 \in [T/2, T)$, and has 0 for other entries.
RNNs are required to output a target value $a_{i_1} + a_{i_2}$ after reading all tokens.
We trained and evaluated models on this task of sequence length $T = 5000$ with a mean squared error (MSE).

\textbf{Copy task.}
The input sequence of length $T +20$ consists of 10 alphabets.
The first ten are taken randomly from eight alphabets $\{a_1,\cdots,a_8\}$, and others are all \emph{void} token $a_0$ except the $(T+11)$-th \emph{revoking} token $a_9$. 
RNNs are required to output the first ten tokens after reading the revoking input.
Each alphabet is encoded as a one-hot vector. 
The models were trained on this task of void sequence length $T=500$ with the cross-entropy loss calculated over the last 10 tokens as in a previous study~\cite{gua2020improving} and evaluated in terms of accuracy.

On both tasks, we generated 64 training data randomly at each training iteration instead of creating a dataset beforehand.
We set the hidden dimension to 128 for each model, as in a previous study~\cite{gua2020improving}.
For NRU, the memory cell dimension was set to 64 following the original paper~\cite{chandar2019towards}.
We used RMSProp with a learning rate $10^{-3}$ and other parameters set as default ($\alpha=0.99, \epsilon=10^{-8}$) to train each model.
We used the gradient clipping~\cite{pascanu2013difficulty} with threshold $1$ for stable training.
For the observation of the growth of time scales of models (\rfig{fig:adding_timescale}), models were trained in the same setting above except that the learning rate was set to $10^{-2}$ to see clearer difference.

\subsection{Pixel-by-pixel Image Recognition}\label{app:image_classification}

We generally follow the way of \citet{gua2020improving} to train models on image recognition tasks.
In the psMNIST task, the means of permutation of pixels
adopts bit-reversal permutation rather than random permutation.
In the sCIFAR task, recurrent models are given a three-dimensional pixel value at each time step, so the sequence length is 1024 time steps.
The hidden dimension of each model was set to 512 on all tasks.
We used Adam~\cite{kingma2014adam} for training and the learning rate was swept over $\{0.0005, 0.0002\}$.
The best stable learning rate was $0.0002$ for the LSTM with the softsign gate and the NRU and $0.0005$ for the other models on every tasks.
The gradient clipping~\cite{pascanu2013difficulty} was applied with threshold 1.
We trained models with 150 epochs with the batch size 50.
The memory size for the NRU is set to 144 as in~\citet{chandar2019towards}.

\subsection{Language Modeling}

We explain the implementation details of the language modeling experiment.
We used the public code of \citet{mahto2020multi} to train and evaluate the language models and used the same hyperparameters.
However, their code causes an out-of-memory error of GPUs during training in our computational environment.
The memory usage increases after switching the optimizer to ASGD, then training breaks after 324 epochs.
We completed the training by successively retraining each model twice after a break in training due to the out-of-memory error.
Furthermore, each LSTM module was manually re-implemented on the basis of pytorch library without low-level optimization.
Although this results in longer runtime for the baseline LSTM than the original pytorch implementation of LSTM, 
we did this to fairly compare the runtime.


\section{Additional Experimental Results}\label{app:additional_experiments}

\begin{figure}[t]
    \centering
    \includegraphics[width=1.0\linewidth]{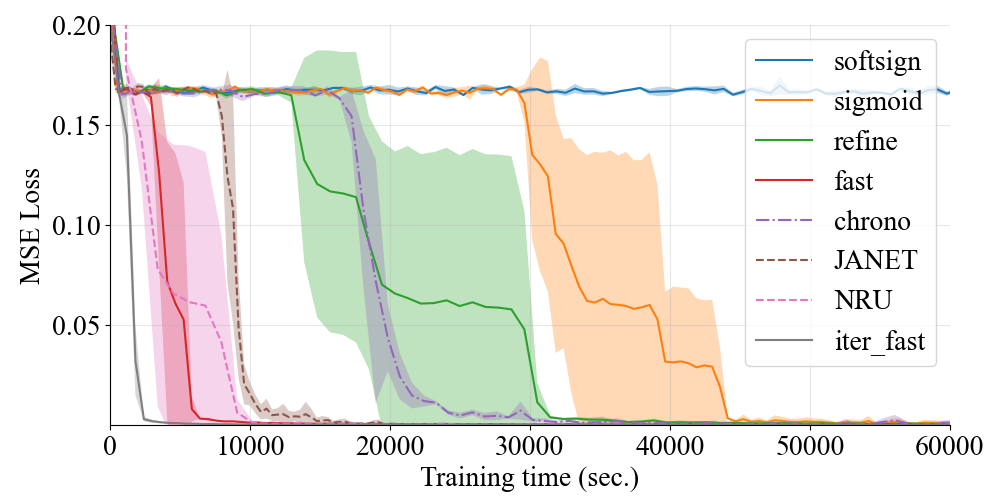}
    \caption{
        MSE for adding task of length 5000 vs elapsed time
    }
    \label{fig:adding_runtime}
\end{figure}

\begin{figure}[t]
    \centering
    \includegraphics[width=\linewidth]{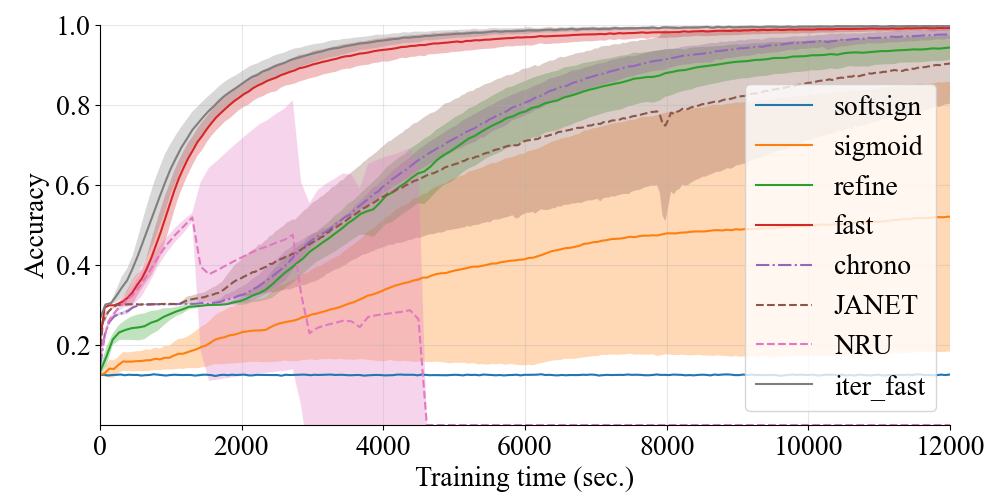}
    \caption{
        Accuracy for copy task of length 500 vs elapsed time
    }
    \label{fig:copy_acc_runtime}
\end{figure}

\begin{figure}[t]
    \centering
    \includegraphics[width=\linewidth]{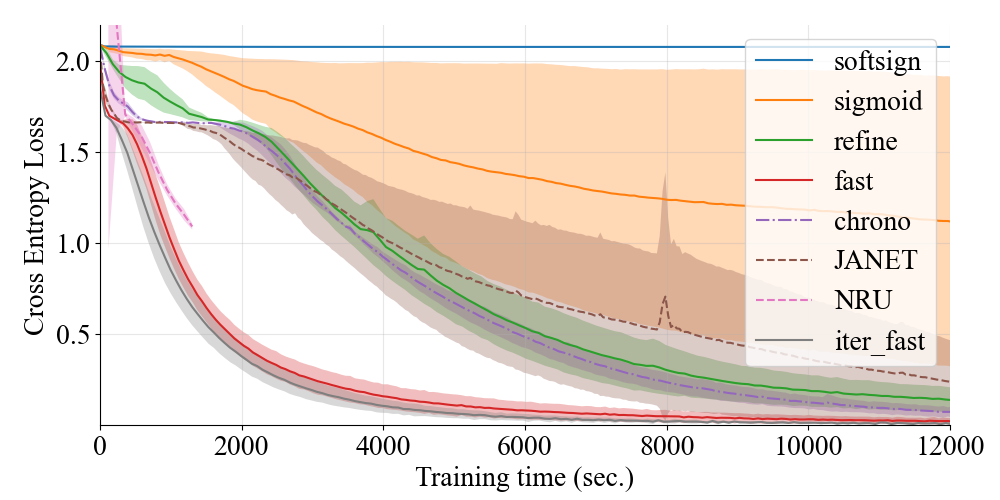}
    \caption{
        Cross entropy loss for copy task of length 500 vs elapsed time
    }
    \label{fig:copy_loss}
\end{figure}

\begin{table*}[t]
    \centering
    \begin{tabular}{rcccc}
        \hline
        & sMNIST & psMNIST & sCIFAR & Time / Epoch \\
        \hline
        Softsign LSTM     & 21.63 $\pm$ 10.15 & 91.71 $\pm$ 0.33 & 59.21 $\pm$ 0.39 & 17.7 min.\\
        Sigmoid LSTM      & 98.88 $\pm$ 0.12  & 95.71 $\pm$ 0.02 & 69.14 $\pm$ 0.39 & 14.3 min.\\
        Refine LSTM       & 98.94 $\pm$ 0.03  & 95.93 $\pm$ 0.16 & 69.55 $\pm$ 0.50 & 22.7 min.\\
        UR-LSTM       & 98.80 $\pm$ 0.14  & 96.11 $\pm$ 0.08 & 70.29 $\pm$ 0.12 & 21.7 min. \\
        \textbf{Fast LSTM} & 99.05 $\pm$ 0.04  & 96.18 $\pm$ 0.14 & 70.06 $\pm$ 0.38 & 14.7 min.\\
        \textbf{UF-LSTM}   & 99.01 $\pm$ 0.01  & 96.11 $\pm$ 0.03 & 70.04 $\pm$ 0.33 & 14.7 min. \\
        Iterated Fast LSTM & 99.02 $\pm$ 0.07 & 96.22 $\pm$ 0.08 & 69.66 $\pm$ 0.41 & 15.8 min. \\
        \hline
        Softsign GRU     & 69.71 $\pm$ 41.27 & 93.80 $\pm$ 0.28 & 61.90 $\pm$ 1.33 & 17.5 min. \\
        Sigmoid GRU      & 98.99 $\pm$ 0.06  & 95.23 $\pm$ 0.07 & 68.92 $\pm$ 0.48 & 14.9 min. \\
        Refine GRU       & 98.95 $\pm$ 0.02  & 95.61 $\pm$ 0.02 & 69.54 $\pm$ 0.71 & 22.7 min. \\
        UR-GRU        & 98.86 $\pm$ 0.08  & 95.49 $\pm$ 0.13 & 70.29 $\pm$ 0.35 & 22.8 min. \\
        \textbf{Fast GRU} & 98.95 $\pm$ 0.05  & 95.54 $\pm$ 0.34 & 69.65 $\pm$ 0.18 & 15.5 min.\\
        \textbf{UF-GRU}   & 98.79 $\pm$ 0.02  & 95.46 $\pm$ 0.30 & 69.59 $\pm$ 0.34 & 15.9 min. \\
        \hline
    \end{tabular}
    \caption{Test accuracy on image classification tasks with various LSTM and GRU. UR- and UF- stands for refine and fast gate with uniform initialization~\cite{gua2020improving} of forget gate bias. Processing time per epoch on psMNIST is reported.}
    \label{tab:sMNIST_results_GRU}
\end{table*}

For completeness, we further demonstrated the effectiveness of the fast gate on additional experiments.

\subsection{Synthetic Tasks Evaluated on Elapsed Time}
\label{app:exp_runtime}
To show the efficiency of learning with the fast gate based on the processing time rather than the number of parameter updates,
we plot the mean squared error on the adding task to wall clock training time on \rfig{fig:adding_runtime}.
We observe that the fast gate achieves the convergence speed comparable to NRU due to less computational overhead on the state update.
We also show results on accuracy and cross entropy loss for the copy task based on wall clock time in \rfig{fig:copy_acc_runtime} and \rfig{fig:copy_loss}.
The learning curve of NRU for the loss is truncated since its training diverges.
Training with the fast gate converges the fastest among all baselines. As explained in Appendix~\ref{app:choice_of_gate}, the iterated fast gate $\phi(z) = \sigma(\sinh (\sinh z))$ converges even faster than the fast gate on both tasks.

\subsection{Fast gate combined with other methods}
\label{app:exp_gru}
Since our method is in principle applicable to \emph{any} gated RNNs, we further validate the fast gate on another popular gated RNN, GRU~\cite{cho2014learning}.
Additionally, we evaluate the fast gate combined with the gate initialization technique, called uniform initialization~\cite{gua2020improving}. 

In \rtab{tab:sMNIST_results_GRU}, we see similar (but slightly smaller) improvement of accuracy of the GRU in accuracy as that of the LSTM on the psMNIST and sCIFAR tasks, while keeping the computational cost.
There is almost no difference in accuracy on the sMNIST task among the sigmoid, refine and fast gates because sMNIST contains relatively short-term dependencies.
For the fast gate, combination of uniform initialization does not contribute to further performance improvement unlike the refine gate.
Note that for the refine gate, the uniform initialization is applied to both the original gate and the auxiliary gate~\cite[Appendix~A]{gua2020improving}, 
which actually provides different time scale distribution between the refine gate and other gates.
That is, the time scale distribution of the refine gate with uniform initialization results in \emph{skewed} distribution different from $\mathbb P(D=x) \propto 1/x^2$ unlike explained in~\cite[Proposition~1]{gua2020improving}.
We suspect that this skewness leads to better improvements of the refine gate combined the uniform initialization.
More sophisticated initialization methods suited also to other gates are to be explored as future work.

\subsection{Image classification with limited model size}
\label{app:image_experiment_small}

\begin{table}[t]
    \centering
    \resizebox{\linewidth}{!}{
    \begin{tabular}{ccccc}
        \hline
        & sMNIST & psMNIST & sCIFAR & Time \\
        \hline 
        Softsign      & 97.20 $\pm$ 0.20  & 87.89 $\pm$ 0.76 & 47.91 $\pm$ 1.13 & 345 sec.\\
        Sigmoid       & 98.84 $\pm$ 0.08  & 92.80 $\pm$ 0.37 & 62.73 $\pm$ 0.75 & 284 sec.\\
        Refine        & 98.88 $\pm$ 0.12  & 93.28 $\pm$ 0.27 & 62.10 $\pm$ 0.68 & 486 sec.\\
        UR-       & 98.69 $\pm$ 0.08  & 92.56 $\pm$ 0.46 & 62.54 $\pm$ 0.85 & 484 sec. \\
        \textbf{Fast} & 98.81 $\pm$ 0.08  & 93.83 $\pm$ 0.15 & 62.98 $\pm$ 0.29 & 314 sec.\\
        \textbf{UF-}  & 98.81 $\pm$ 0.07  & 93.00 $\pm$ 0.54 & 63.00 $\pm$ 0.78 & 315 sec.\\
        \hline
        chrono   & 97.96 $\pm$ 0.16  & 90.32 $\pm$ 0.62 & 54.71 $\pm$ 0.50 & 287 sec. \\
        JANET         & 98.07 $\pm$ 0.09 & 87.84 $\pm$ 0.50 & 55.82 $\pm$ 0.35 & 214 sec.\\
        NRU           & 98.19 $\pm$ 0.22 & 94.89 $\pm$ 0.15 & 53.34 $\pm$ 1.06 & 802 sec.\\ %
        \hline
    \end{tabular}
    }
    \caption{Test accuracy on image classification tasks with limited model size on four random seeds. Gate variants are applied to LSTM. UR-/UF- is refine and fast gate combined with uniform initialization. Processing time per epoch on psMNIST is reported.}
    \label{tab:sMNIST_small_results}
\end{table}

As pointed out in~\cite{voelker2019legendre}, improvements in learning long term dependencies could be better compared under the limits of model size.
Therefore, we also tested a smaller classifier for sMNIST, psMNIST and sCIFAR than that in the main text in terms of the number of the output layer (1 vs 2) and hidden dimension of the recurrent layer (128 vs 512), which is commonly adopted in the literature~\cite{chandar2019towards,chang2019antisymmetricrnn}.
The memory size of NRU is set to 64 in this setting as in the original paper.

The results in \rtab{tab:sMNIST_small_results} are generally consistent with the main text, 
which shows robustness of the performance improvement of the fast gate.
While the NRU shows the highest accuracy in the psMNIST task, it does not scale to larger models in terms of accuracy and the processing time as in the main text.
For training, we used RMSprop with the learning rate 0.0005 except the following.
On the sMNIST and sCIFAR tasks, learning rate of 0.0002 is applied to the NRU since it suffers from training instability at higher learning rates.
The gradient clipping~\cite{pascanu2013difficulty} was applied with threshold 1.
We trained models with 150 epochs on these task, and 100 epochs on the sCIFAR task.
The learning rate was divided by 5 after 100 epochs on the sMNIST and psMNIST tasks, and after 50 epochs on the sCIFAR task.
The batch size was set to 100 on all tasks.
The models are trained with four different random seeds.

\subsection{Raw speech classification}
\label{app:raw-speech}

\begin{table}[t]
    \centering
    \begin{tabular}{rc}
        \hline
        Model & Test Accuracy \\
        \hline
        Sigmoid LSTM         &  21.5* \\ 
        Refine LSTM  &  83.9* \\
        UR-LSTM      &  77.6* \\
        \textbf{Fast LSTM} & 92.3 \  \\
        \hline
    \end{tabular}
    \caption{Test accuracy on \emph{raw} Speech Command (SC) dataset \cite{romero2021ckconv}. 
    Asterisks (*) denotes divergence of training.
    }
    \label{tab:results_sc}
\end{table}

To see effectiveness of the fast gate on much longer sequences, we trained LSTMs with various gates on raw Speech Command dataset~\cite{romero2021ckconv}, which consists of sequences of length 16,000.
We adopted the public code\footnote{https://github.com/HazyResearch/state-spaces} to train 1-layer LSTM of hidden dimension 256 for classifying speech data into 10 classes.
We search the best learning rate from $\{1\times 10^{-3}, 5 \times 10^{-4}, 1 \times 10^{-4}, 5 \times 10^{-5}\}$ and gradient clipping threshold from $\{ 1, 0.1\}$.

The results in \rtab{tab:results_sc} show that gates other than the fast gate suffer from training instability.
In particular, the model with the sigmoid gate diverges with few epochs.
We see that the refine gate improves learning, but the training diverges before it reaches sufficiently high accuracy.
On the other hand, the LSTM with the fast gate can stably learn and achieve the best accuracy.
These results indicate that a gate function with higher order saturation enables more stable training on such extremely long sequences.

\end{document}